\documentclass[10pt,twocolumn,letterpaper]{article}

\usepackage[pagenumbers]{cvpr}
\usepackage{cvpr}
\usepackage{graphicx}
\usepackage{amsmath}
\usepackage{amssymb}
\usepackage{booktabs}
\usepackage{multirow}
\usepackage{bbm}
\usepackage{pifont}
\usepackage[ruled,vlined, noend]{algorithm2e}
\usepackage[dvipsnames]{xcolor}
\definecolor{citecolor}{HTML}{0071bc}

\usepackage[pagebackref,breaklinks,colorlinks, anchorcolor=blue, citecolor=citecolor]{hyperref}

\usepackage[capitalize]{cleveref}
\crefname{section}{Sec.}{Secs.}
\Crefname{section}{Section}{Sections}
\Crefname{table}{Table}{Tables}
\crefname{table}{Tab.}{Tabs.}

\newcommand{\cmark}{\ding{51}}%
\newlength\savewidth

\def\cvprPaperID{6256}
\def\confName{CVPR}
\def\confYear{2022}

\begin{document}

\title{ST++: Make Self-training Work Better for Semi-supervised Semantic Segmentation}

\author{Lihe Yang\textsuperscript{\rm 1}~~~~~ Wei Zhuo\textsuperscript{\rm 3}~~~~~ Lei Qi\textsuperscript{\rm 4,1}~~~~~ Yinghuan Shi\textsuperscript{\rm 1,2}\thanks{Corresponding author. Work supported by National Key Research and Development Program of China (2019YFC0118300), NSFC Major Program (62192783), CAAI-Huawei MindSpore Project (CAAIXSJLJJ-2021-042A), China Postdoctoral Science Foundation Project (2021M690609), and Jiangsu Natural Science Foundation Project (BK20210224).}~~~~~ Yang Gao\textsuperscript{\rm 1}\vspace{1mm}\\
\textsuperscript{\rm 1}State Key Laboratory for Novel Software Technology, Nanjing University\\
\textsuperscript{\rm 2}National Institute of Healthcare Data Science, Nanjing University\\
\textsuperscript{\rm 3}Tencent\hspace{8mm}\textsuperscript{\rm 4}Southeast University\\
\small \texttt{lihe.yang.cs@gmail.com~~ weizhuo@tencent.com~~ qilei@seu.edu.cn~~ \{syh, gaoy\}@nju.edu.cn}}

\maketitle

\begin{abstract}

Self-training via pseudo labeling is a conventional, simple, and popular pipeline to leverage unlabeled data.
In this work, we first construct a strong baseline of self-training (namely ST) for semi-supervised semantic segmentation via injecting strong data augmentations (SDA) on unlabeled images to alleviate overfitting noisy labels as well as decouple similar predictions between the teacher and student. With this simple mechanism, our ST outperforms all existing methods without any bells and whistles, e.g., iterative re-training. Inspired by the impressive results, we thoroughly investigate the SDA and provide some empirical analysis.
Nevertheless, incorrect pseudo labels are still prone to accumulate and degrade the performance.
To this end, we further propose an advanced self-training framework (namely ST++), that performs selective re-training via prioritizing reliable unlabeled images based on holistic prediction-level stability. 
Concretely, several model checkpoints are saved in the first stage supervised training, and the discrepancy of their predictions on the unlabeled image serves as a measurement for reliability. Our image-level selection offers holistic contextual information for learning. We demonstrate that it is more suitable for segmentation than common pixel-wise selection. As a result, ST++ further boosts the performance of our ST. Code is available at \url{https://github.com/LiheYoung/ST-PlusPlus}.

\end{abstract}
\section{Introduction}

Fully-supervised semantic segmentation learns to assign pixel-wise semantic labels via generalizing from numerous densely annotated images. Despite the rapid progress \cite{zhao2017pyramid, chen2018encoder}, the pixel-wise manual labeling is costly, laborious, and even infeasible, precluding its deployment in some scenes such as medical image analysis. To avert the labor-intensive procedure, semi-supervised semantic segmentation has been proposed to learn a model from a handful of labeled images along with abundant unlabeled images.

\begin{figure}
    \centering
    \includegraphics[width=0.95\linewidth]{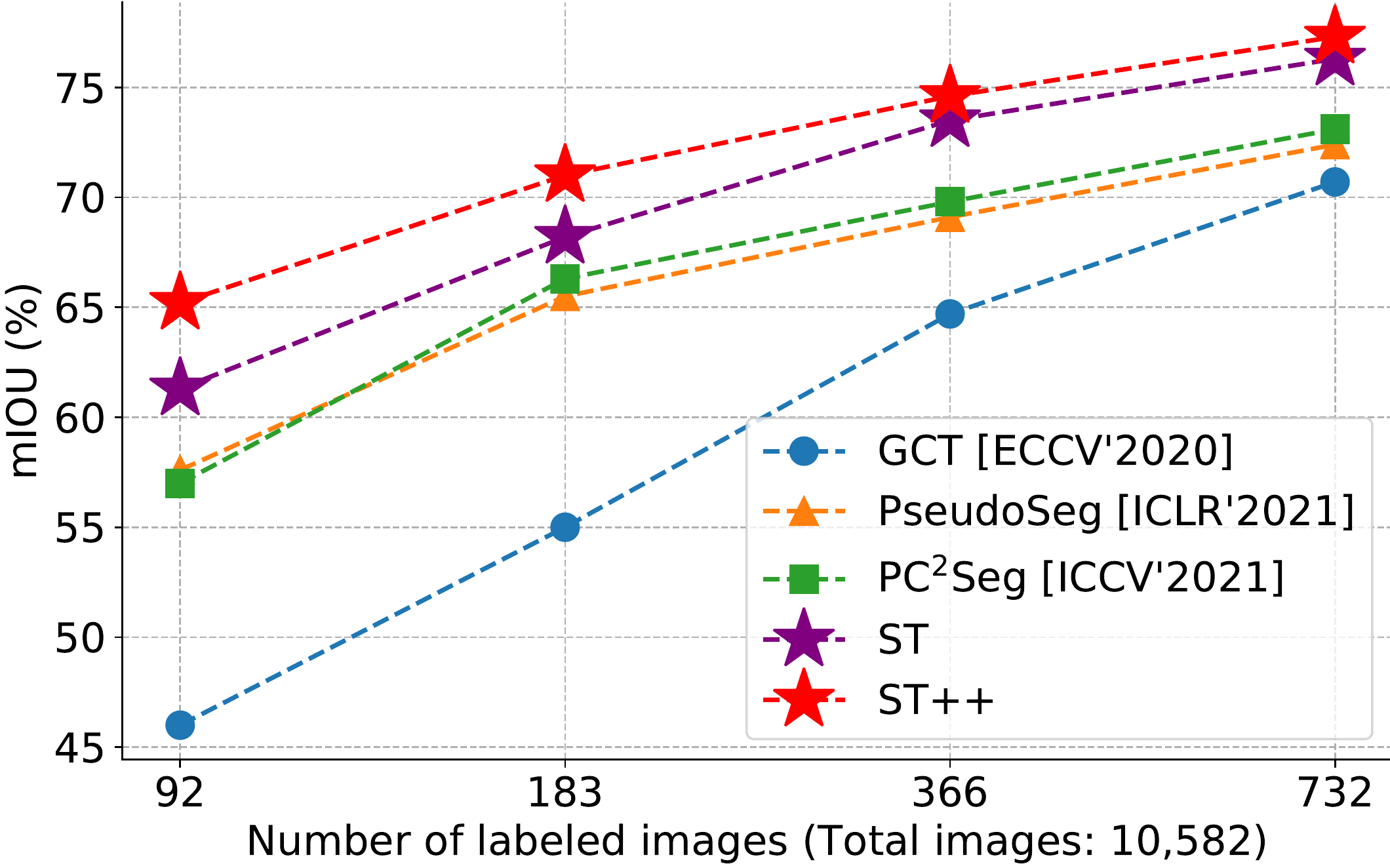}
    \caption{Performance comparisons between our ST/ST++ and the state-of-the-art methods on the Pascal. It is worth noting that the proposed ST and ST++ surpass previous best results significantly, especially in the extremely scarce-label regime, \eg, 92 labels.}
    \vspace{-0.45cm}
    \label{fig:intro}
\end{figure}

The core challenge in semi-supervised setting lies in how to effectively utilize the unlabeled images. Prior works in semi-supervised learning (SSL) propose to apply entropy minimization \cite{lee2013pseudo, xie2020self} or consistency regularization \cite{grandvalet2005semi, xie2019unsupervised} on unlabeled images. With increasingly sophisticated mechanisms introduced to this field, FixMatch \cite{sohn2020fixmatch} breaks the trend and achieves inspiring results via integrating both strategies into a hybrid framework with few hyper-parameters. Motivated by the tremendous progress in SSL, recent works in semi-supervised semantic segmentation have evolved from GANs-based methods \cite{goodfellow2014generative, mittal2019semi} to delving into consistency regularization from the segmentation perspective, such as enforcing consistent predictions of the same unlabeled image under strong-weak perturbations \cite{zou2020pseudoseg}, of the same local patch from different contextual crops \cite{lai2021semi}, and of the same unlabeled image between dual differently initialized models \cite{feng2020dmt, chen2021semi}.

\emph{Nevertheless, are the delicate mechanisms indispensable for semi-supervised semantic segmentation? More importantly, is the straightforward self-training scheme \cite{lee2013pseudo} proposed around a decade ago already out-of-date for this task?} In this work, we intuitively and empirically present two simple and effective techniques to bring back the classical self-training method as a strong competitor again.

The self-training \cite{lee2013pseudo} is commonly regarded as a form of entropy minimization in SSL, since the re-trained student is supervised with hard labels produced by the teacher which is trained on labeled data. However, they suffer severe coupling issue, \ie, making similar predictions on the same input. We notice that, however, injecting strong data augmentations (SDA) on unlabeled images is extremely beneficial to decouple their predictions as well as alleviate overfitting on noisy pseudo labels. 
Despite the simplicity, self-training with SDA significantly surpasses existing methods without any bells and whistles, \eg, without the need of iterative re-training \cite{yuan2021simple}, manually choosing a threshold for filtering incorrect labels \cite{zou2020pseudoseg}, or repetitively producing pseudo labels for each training minibatch \cite{zou2020pseudoseg, chen2021semi, hu2021semi}. Inspired by the impressive results, we thoroughly investigate the SDA and find that it is fully compatible with the off-the-shelf augmentation strategies in contrastive learning \cite{chen2020simple}, \eg, colorjitter, grayscale, and blur, which deteriorate clean data distribution but perform surprisingly well for unlabeled data. Besides, we examine individual effectiveness of each data augmentation, and observe that the simple colorjitter plays most effectively and different augmentations are complementary to each other. Formally, this basic self-training framework with SDA is named as ST in this work, serving as a strong baseline for our full method.

Another longstanding but underestimated issue is that the classical self-training framework utilizes all unlabeled images at the same time. Nevertheless, different unlabeled images cannot be equally easy \cite{ren2020not, li2017not, sun2019not} and the corresponding pseudo labels cannot be equally reliable, leading to severe confirmation bias \cite{arazo2020pseudo} and potential performance degradation when iteratively optimizing the model with those ill-posed pseudo labels. To this end, we further propose an advanced ST++ framework based on our ST, that automatically selects and prioritizes more reliable images in the re-training phase to produce higher-quality artificial labels on the remaining less reliable images. The measurement for the reliability or uncertainty of an unlabeled image is to compute the holistic stability of the evolving pseudo masks in different iterations during the entire training course. Note that, different from the common practice of manually setting a fixed confidence threshold to filter low-confidence pixels \cite{zou2020pseudoseg}, we demonstrate that our image-level selection based on the stability of evolving predictions can provide holistic contextual regions for model training, which is more appropriate to the segmentation task.

It is worth noting that the classical self-training pipeline \cite{lee2013pseudo} is attracting increasing attention \cite{radosavovic2018data, xie2020self} in the semi-supervised setting. Our work differs from them in that we empirically and systematically study the effectiveness of strong data augmentations on unlabeled data and further propose an advanced self-training framework with selective re-training property. More concrete differences are discussed in detail in the related work. Our main findings and contributions are summarized as follows:
\vspace{-0.15cm}

\begin{itemize}
    \item We construct a strong baseline (ST) of self-training in semi-supervised semantic segmentation via injecting strong data augmentations on unlabeled images during re-training. Motivated by the promising performance, we provide intuitive explanations and systematically investigate the role of SDA and each augmentation.
    \vspace{-0.15cm}
    
    \item Built on our ST, to alleviate the potential performance degradation incurred by incorrect pseudo labels, we further propose an advanced self-training framework ST++, that performs selective re-training via prioritizing reliable images based on holistic prediction-level stability in the entire training course. We demonstrate that the image-level selection is more suitable for segmentation task compared with pixel-wise selection.
    \vspace{-0.15cm}
    
    \item The ST and ST++ both outperform previous methods across extensive settings and architectures on the Pascal and Cityscapes dataset, with few hyper-parameters.
\end{itemize}

\section{Related Work}

\noindent
\textbf{Semi-supervised learning.} Two main branches of methods are proposed in recent years, namely consistency regularization \cite{grandvalet2005semi, bachman2014learning, sajjadi2016regularization, laine2016temporal, tarvainen2017mean, miyato2018virtual, xie2019unsupervised} and entropy minimization \cite{lee2013pseudo, xie2020self, cascante2020curriculum, pham2021meta}. Consistency regularization enforces the current optimized model to yield stable and consistent predictions under various perturbations \cite{sajjadi2016regularization, xie2019unsupervised}, \eg, shape and color, on the same unlabeled data. Earlier works also save several checkpoints \cite{laine2016temporal} or maintain a teacher whose parameters are the exponential moving average of the updating student \cite{tarvainen2017mean} to produce more reliable artificial labels for student model. On the other hand, entropy minimization, popularized by the self-training pipeline \cite{grandvalet2005semi, lee2013pseudo}, leverages unlabeled data in an explicit bootstrapping manner, where unlabeled data is assigned with pseudo labels to be jointly trained with manually labeled data. Different from prior works, MixMatch \cite{berthelot2019mixmatch} harvests advantages of both methodologies and proposes a hybrid framework to exploit the unlabeled data from the two perspectives. FixMatch \cite{sohn2020fixmatch} inherits the spirit from MixMatch but simplifies unnecessary mechanisms. As a further extension, the most recent work FlexMatch \cite{zhang2021flexmatch} utilizes the inherent learning status to filter low-confidence labels with class-wise thresholds.

\vspace{0.1cm}
\noindent
\textbf{Semi-supervised semantic segmentation.} Slightly different from the trend in SSL, preliminary works \cite{souly2017semi, hung2018adversarial, mittal2019semi} in semi-supervised semantic segmentation tend to utilize the Generative Adversarial Networks (GANs) \cite{goodfellow2014generative} as an auxiliary supervision signal for the unlabeled data. However, GANs are not easy to optimize and may suffer the problem of mode collapse \cite{salimans2016improved}. Therefore, also inspired by the success in SSL, subsequent methods \cite{french2019semi, ouali2020semi, zou2020pseudoseg, lai2021semi, chen2021semi, yuan2021simple, zhou2021c3, he2021re, zhong2021pixel, alonso2021semi, hu2021semi, zhang2021robust} manage to tackle this task with simpler mechanisms, such as enforcing similar predictions under multiple perturbed embeddings \cite{ouali2020semi}, under two different contextual crops \cite{lai2021semi}, and between dual differently initialized models \cite{chen2021semi}. As an extension of FixMatch \cite{sohn2020fixmatch}, PseudoSeg \cite{zou2020pseudoseg} adapts the weak-to-strong consistency to segmentation scenario and further applies a calibration module to refine the pseudo masks. Despite fancy mechanisms proposed and rapid progress made, nevertheless, we hope to raise a new observation to this field that, the plainest self-training framework coupled with strong data augmentations (SDA) is indeed effective enough to obtain state-of-the-art performance without any bells and whistles, \eg, iterative re-training or setting a threshold to filter unreliable pixels.

\setlength{\textfloatsep}{0.2cm}
\begin{algorithm}[t]
\linespread{1.0}\selectfont
\caption{\label{alg:st}ST Pseudocode}

\SetAlgoNoLine

\KwIn{Labeled training set $\mathcal{D}^l = \{(x_i, y_i)\}_{i=1}^{M}$, 

~~~~~~~~~~~~Unlabeled training set $\mathcal{D}^u = \{u_i\}_{i=1}^{N}$,

~~~~~~~~~~~~Weak/strong augmentations $\mathcal{A}^w$/$\mathcal{A}^s$,

~~~~~~~~~~~~Teacher/student model $T/S$}
\vspace{0.05cm}
\KwOut{Fully trained student model $S$}
\vspace{0.1cm}
Train $T$ on $\mathcal{D}^l$ with cross-entropy loss $\mathcal{L}_{ce}$

Obtain pseudo labeled $\hat{\mathcal{D}}^u = \{(u_i, T(u_i))\}_{i=1}^N$ 

Over-sample $\mathcal{D}^l$ to around the size of $\hat{\mathcal{D}}^u$

\For {minibatch $\{(x_k, y_k)\}_{k=1}^B \subset (\mathcal{D}^l \cup \hat{\mathcal{D}}^u)$} {

\For {$k \in \{1, \dots, B\}$} {
    \eIf {$x_k \in \mathcal{D}^u$} {
        $x_k, y_k \leftarrow $ $\mathcal{A}^s(\mathcal{A}^w((x_k, y_k))$
    } {
        $x_k, y_k \leftarrow $ $\mathcal{A}^w(x_k, y_k)$
    }
    
    $\hat{y}_k = S(x_k)$
}

Update $S$ to minimize $\mathcal{L}_{ce}$ of $\{(\hat{y}_k, y_k)\}_{k=1}^B$
}

\Return {$S$}
\end{algorithm}

\vspace{0.05cm}
\noindent
\textbf{Self-training.} The self-training via pseudo labeling is an explicit and classical method originating from around a decade ago \cite{lee2013pseudo}. Recently, it is increasingly attracting attention from multiple fields, such as fully-supervised image recognition \cite{radosavovic2018data, yalniz2019billion, xie2020self, zoph2020rethinking}, semi-supervised learning \cite{sohn2020simple, cascante2020curriculum, feng2020dmt, yuan2021simple}, and domain adaptation \cite{zou2018unsupervised, zou2019confidence, kumar2020understanding}. In the semi-supervised setting, particularly, it has been revisited in several tasks, including image classification \cite{cascante2020curriculum}, object detection \cite{sohn2020simple}, and semantic segmentation \cite{yuan2021simple, feng2020dmt}. Among them, the most related ones to us are \cite{yuan2021simple, sohn2020simple}. Nevertheless, our work is fundamentally different from \cite{yuan2021simple} in that we demonstrate appropriate SDA on unlabeled data are extremely beneficial to the semi-supervised learner, while \cite{yuan2021simple} designs their method based on the assumption that excessive data augmentations are destructive to clean data distribution. Another work \cite{sohn2020simple} addresses object detection task via manually designing task-relevant augmentations, whereas our SDA is common in image recognition but previously neglected in semi-supervised segmentation. Moreover, both aforementioned works adopt the plain training pipeline, whereas we further propose the ST++ to safely exploit unlabeled images in a curriculum learning manner \cite{bengio2009curriculum}.

\vspace{0.1cm}
\noindent
\textbf{Uncertainty estimation.} Previous method \cite{kendall2017uncertainties} estimates model uncertainty with a Bayesian analysis. However, limited by the computational burden of Bayesian inference, some other methods adopt Dropout \cite{gal2016dropout, liang2021r} and data augmentations \cite{ayhan2018test} to measure the uncertainty. In the semi-supervised setting, FixMatch \cite{sohn2020fixmatch} simply sets a confidence threshold to filter uncertain samples, and DMT \cite{feng2020dmt} maintains two differently initialized networks to highlight disagreed regions. Compared with them, our method estimates image-level uncertainty via measuring the holistic prediction stability of evolving masks without the need of training extra networks or manually choosing the threshold, making it universal to other scenes. Also, the model learns holistic contextual regions in high-confidence images, which is more stable and appropriate to the segmentation task.

\begin{algorithm}[t]
\linespread{1.16}\selectfont
\caption{\label{alg:st++}ST++ Pseudocode}
\SetAlgoNoLine

\vspace{0.06cm}
\KwIn{Same as Algorithm~\ref{alg:st}}
\KwOut{Same as Algorithm~\ref{alg:st}}

\vspace{0.1cm}

Train $T$ on $\mathcal{D}^l$ and save $K$ checkpoints $\{T_j\}_{j=1}^K$

\For {$u_i \in \mathcal{D}^u$} {
    \For {$T_j \in  \{T_j\}_{j=1}^K$} {
    \vspace{0.06cm}
        Pseudo mask $M_{ij} = T_j(u_i)$
    }
    Compute $s_i$ with Equation \ref{eq:s} and $\{M_{ij}\}_{j=1}^K$
}
\vspace{0.02cm}
Select $R$ highest scored images to compose $\mathcal{D}^{u_1}$

$\mathcal{D}^{u_2} = \mathcal{D}^u - \mathcal{D}^{u_1}$

$\mathcal{D}^{u_1} = \{(u_k, T(u_k))\}_{u_k \in \mathcal{D}^{u_1}}$

Train $S$ on $(\mathcal{D}^l \cup \mathcal{D}^{u_1})$ with ST re-training

$\mathcal{D}^{u_2} = \{(u_k, S(u_k))\}_{u_k \in \mathcal{D}^{u_2}}$

Re-initialize $S$

Train $S$ on $(\mathcal{D}^l \cup \mathcal{D}^{u_1} \cup \mathcal{D}^{u_2})$ with ST re-training

\Return {$S$}
\end{algorithm}

\section{Method}

\subsection{Problem Definition}
Semi-supervised semantic segmentation aims to generalize from a combination set of pixel-wise labeled images $\mathcal{D}^l = \{(x_i, y_i)\}_{i=1}^M$ and unlabeled images $\mathcal{D}^u = \{u_i\}_{i=1}^N$, where in most cases $N \gg M$. In most works, the overall optimization target is formalized as:
\vspace{-0.1cm}
\begin{equation}
    \mathcal{L} = \mathcal{L}^s + \lambda\mathcal{L}^u,
\vspace{-0.1cm}
\end{equation}
where $\lambda$ acts as a tradeoff between labeled and unlabeled data. It can be a fixed value \cite{sohn2020fixmatch, sohn2020simple} or be scheduled during training \cite{jeong2019consistency}. The unsupervised loss $\mathcal{L}^u$ is the key point to distinguish different semi-supervised methods, while the supervised loss $\mathcal{L}^s$ is typically the cross-entropy loss between predictions and manually annotated masks.

\subsection{Plainest Self-training Scheme}

We simplify the plainest form of self-training from \cite{lee2013pseudo}. It includes three steps without the need of iterative training:
\begin{enumerate}
\vspace{-0.15cm}
    \item \lbrack Supervised Learning\rbrack~Train a teacher model $T$ on $\mathcal{D}^l$ with cross-entropy loss.
    \vspace{-0.15cm}
    
    \item \lbrack Pseudo Labeling\rbrack~Predict one-hot hard pseudo labels on $\mathcal{D}^u$ with $T$ to obtain $\hat{\mathcal{D}}^u = \{(u_i, T(u_i))\}_{i=1}^N$.
    \vspace{-0.15cm}
    
    \item \lbrack Re-training\rbrack~Re-train a student model $S$ on the union set $\mathcal{D}^l \cup \hat{\mathcal{D}}^u$ for final evaluation.
    \vspace{-0.15cm}
\end{enumerate}

Here, the unsupervised loss $\mathcal{L}^u$ can be formulated as:
\begin{equation}
    \mathcal{L}^u_{plain} = \textrm{H}\big(T(x), S(\mathcal{A}^w(x))\big),
\end{equation}
where $T$ and $S$ map the image $x$ to the output space. $\mathcal{A}^w$ applies random weak data augmentations to the raw image. H denotes entropy minimization between student and teacher.

\vspace{0.1cm}
\noindent
\textbf{Discussion.} Since self-training has largely lagged behind consistency based methods in SSL \cite{sohn2020fixmatch}, we provide intuitive explanations for the promising performance it might achieve in semantic segmentation. Self-training is deemed to heavily rely on the initial model trained with labeled data, which cannot be well satisfied in the scarce-label regime of SSL. However, the situation is different in our task, since all labeled images are densely annotated and supervised, which means that even only tens of labeled images are available, millions of pixel-level samples can be utilized for training, yielding a well-performed model for pseudo labeling.

\subsection{ST: Inject SDA on Unlabeled Images}

The above self-training scheme has long been criticized for its ill-posed property that errors in pseudo labels will accumulate and considerably degrade student performance when iteratively overfitting the incorrect supervision. Moreover, in such a bootstrapping process, inadequate information is introduced during re-training, leading to severe coupling issue between the teacher and re-trained student. Concretely, the re-trained $S$ is enforced to learn the pseudo labels from $T$ in a supervised manner. However, considering the same network structure and similar initialization of $T$ and $S$, they are prone to make similar true or false predictions on the unlabeled images, hence the student $S$ fails to learn extra information except entropy minimization. 

In order to break out of the aforementioned two dilemmas, \ie, overfitting noisy labels and prediction coupling between the student and teacher, we propose to inject strong data augmentations (SDA) on unlabeled images during the re-training phase to pose a more challenging optimization target for the student model. The SDA here is named opposite to the weak or basic augmentations adopted in regular fully-supervised semantic segmentation, including random resizing, random cropping and random flipping. As for specific choices of SDA, we manage to maintain a universal strategy across different datasets or settings, rather than searching for the most appropriate ones in each dataset. To simplify the choice, we adopt the off-the-shelf SDA in \cite{chen2020simple, chen2020improved}, which includes colorjitter, grayscale, and blur. Apart from these color transformations, we introduce another spatial transformation Cutout \cite{devries2017improved} to compose our full SDA. In our ST, the unsupervised objective is formalized as:
\begin{equation}
    \mathcal{L}^u_{ST} = \textrm{H}\big(T(x), S\big(\mathcal{A}^s(\mathcal{A}^w(x))\big)\big),
\end{equation}
where $\mathcal{A}^s$ applies strong data augmentations to the input.

Previous works adjust the impact of labeled and unlabeled data through non-uniform sampling within a minibatch \cite{sohn2020fixmatch} or selecting a hyper-parameter to re-weight the supervised and unsupervised loss \cite{feng2020dmt}. In our ST, we simplify this choice via directly over-sampling $\mathcal{D}^l$ to around the same scale as $\mathcal{D}^u$ and then sampling uniformly from the combined dataset. With this modification, no extra hyper-parameters are introduced and the semi-supervised learner is optimized in a totally fully-supervised fashion. The pseudocode of our ST framework is present in \textbf{Algorithm \ref{alg:st}} and it works as a strong baseline for our full method.

\vspace{0.1cm}
\noindent
\textbf{Discussion.} Despite the simplicity of our ST, it surpasses existing state-of-the-art methods even without iterative re-training \cite{yuan2021simple}. Compared with other online methods \cite{zou2020pseudoseg, chen2021semi, hu2021semi} that repetitively assign pseudo labels for each coming minibatch, our ST annotates unlabeled images only once and the training is conducted in a fully-supervised fashion. Besides, we do not manually fine-tune the choice of SDA. The SDA, harmful to the clean data distribution, but is vital to unlabeled images, please refer to Table \ref{tab:ablation_sda} for detail.

\begin{table*}
    \centering
    \small
    \begin{tabular}{clccc|clccc}
    
    \toprule
    
    \multirow{2}*{Network} & \multirow{2}*{Method} & 1/16 & 1/8 & 1/4 & \multirow{2}*{Network} & \multirow{2}*{Method} & 1/16 & 1/8 & 1/4 \\
    ~ & ~ & (662) & (1323) & (2645) & ~ & ~ & (662) & (1323) & (2645) \\
    \midrule
    
    \multirow{5}*{\shortstack[c]{PSPNet\\\\ ResNet-50}} & SupOnly & 63.8 & 67.2 & 69.6 & \multirow{5}*{\shortstack[c]{DeepLabv3+\\\\ ResNet-50}} & SupOnly & 64.8 & 68.3 & 70.5 \\
    
    ~ & CCT \cite{ouali2020semi} & 62.2 & 68.8 & 71.2 & ~ & ECS \cite{mendel2020semi} & - & 70.2 & 72.6 \\
     
    ~ & DCC \cite{lai2021semi} & 67.1 & 71.3 & 72.5 & ~ & DCC \cite{lai2021semi} & 70.1 & 72.4 & 74.0 \\
    
    ~ & \textbf{ST} & \textbf{\textit{69.1}} & \textbf{\textit{73.0}} & \textbf{\textit{73.2}} & ~ & \textbf{ST} & \textbf{\textit{71.6}} & \textbf{\textit{73.3}} & \textbf{\textit{75.0}} \\
     
    ~ & \textbf{ST++} & \textbf{69.9} & \textbf{73.2} & \textbf{73.4} & ~ & \textbf{ST++} & \textbf{72.6} & \textbf{74.4} & \textbf{75.4} \\
    
    \midrule
    
    \multirow{6}*{\shortstack[c]{DeepLabv2\\\\ ResNet-101}} & SupOnly & 64.3 & 67.6 & 69.5 & \multirow{6}*{\shortstack[c]{DeepLabv3+\\\\ ResNet-101}} & SupOnly & 66.3 & 70.6 & 73.1 \\
    
    ~ & AdvSSL \cite{hung2018adversarial} & 62.6 & 68.4 & 69.9 & ~ & S4GAN \cite{mittal2019semi} & 69.1 & 72.4 & 74.5 \\
    
    ~ & S4L \cite{zhai2019s4l} & 61.8 & 67.2 & 68.4 & ~ & GCT \cite{ke2020guided} & 67.2 & 72.5 & 75.1 \\
    
    ~ & GCT \cite{ke2020guided} & 65.2 & 70.6 & 71.5 & ~ & DCC \cite{lai2021semi} & 72.4 & 74.6 & 76.3 \\
    
    ~ & \textbf{ST} & \textbf{\textit{68.6}} & \textbf{\textit{71.6}} & \textbf{\textit{72.5}} & ~ & \textbf{ST} & \textbf{\textit{72.9}} & \textbf{\textit{75.7}} & \textbf{\textit{76.4}} \\
    
    ~ & \textbf{ST++} & \textbf{69.3} & \textbf{72.0} & \textbf{72.8} & ~ & \textbf{ST++} & \textbf{74.5} & \textbf{76.3} & \textbf{76.6} \\
    
    \bottomrule
    \end{tabular}
    \caption{Results on Pascal VOC. Labeled images are selected from \textbf{augmented} training set. The fraction (\eg, 1/16) and number (\eg, 662) denote the proportion and number of labeled images. SupOnly (supervised baseline): no unlabeled data are leveraged, thus the model is only trained with labeled data. The best results are marked in \textbf{bold}, while the second best ones are in \textbf{\textit{italic bold}}.}
    \vspace{-0.3cm}
    \label{tab:pascal_augmented_sota}
\end{table*}

\subsection{ST++: Select and Prioritize Reliable Images}

Despite the impressive results obtained by the straightforward ST framework, however, it treats each unlabeled sample equally and leverages them in the same way without considering the inherent reliability and difficulty of individual sample. The incorrect predictions in some hard examples may incur negative impact of the training process. Therefore, in current advanced ST++ framework, we further propose a selective re-training scheme via prioritizing reliable unlabeled samples to safely exploit the whole unlabeled set in an easy-to-hard curriculum learning manner \cite{bengio2009curriculum}.

Previous works estimate the reliability or uncertainty of an image or pixel from different perspectives, such as taking the final softmax output as the confidence distribution and filtering low-confidence pixels by pre-defined threshold \cite{zoph2020rethinking, sohn2020fixmatch}, as well as training two differently initialized models to predict the same unlabeled sample and re-weighting the uncertainty-aware loss with their disagreements \cite{feng2020dmt}. In our ST++, we hope to measure the reliability with a single training model without manually choosing the confidence threshold. And for a more stable evaluation of reliability, we filter out unreliable samples based on image-level information rather than widely adopted pixel-level information. The image-level selection also enables the model to learn more holistic contextual patterns during training.

Specifically, we observe that there is a positive correlation between the segmentation performance and the evolving stability of produced pseudo masks during the supervised training phase. Therefore, the more reliable and better predicted unlabeled images can be selected based on their evolving stability during training. More formally, considering an unlabeled image $u_i \in \mathcal{D}^u$, for $K$ checkpoints $\{T_j\}_{j=1}^K$ saved during training, we predict the pseudo masks of $u_i$ with them to obtain $\{M_{ij}\}_{j=1}^K$. Since training model tends to converge and achieve the best performance in the late training stage, we evaluate the meanIOU between each earlier pseudo mask and the final mask. The meanIOU can serve as a measurement for stability and further reflect the reliability of the unlabeled image along with pseudo mask:
\vspace{-0.1cm}
\begin{equation}
    s_i = \sum_{j=1}^{K-1} \texttt{meanIOU}\left(M_{ij}, M_{iK}\right),
    \label{eq:s}
\vspace{-0.1cm}
\end{equation}
where $s_i$ is the stability score, reflecting the reliability of $u_i$.

Obtaining the stability score of all unlabeled images, we sort the whole unlabeled set based on these scores, and select the top $R$ images with the highest scores for the first re-training phase. With a better optimized student model, the remaining unreliable images are re-labeled and the second re-training phase is conducted on the full combination of manually labeled and pseudo labeled data. The pseudocode of our ST++ method is illustrated in \textbf{Algorithm \ref{alg:st++}}.
\section{Experiments}

\subsection{Setup}

\vspace{0.1cm}
\noindent
\textbf{Dataset.} The Pascal VOC 2012 \cite{everingham2015pascal} is composed of 1464 images for training and 1449 images for validation originally. And the training set can be augmented via introducing relatively lower-quality annotations from the SBD dataset \cite{hariharan2011semantic}, resulting in 10582 training images. The Cityscapes \cite{cordts2016cityscapes} contains 2975 images with fine-grained masks for training and 500 images for validation.

\vspace{0.1cm}
\noindent
\textbf{Network structure.} In the past few years, different models and backbones are utilized. In order to conduct a comprehensive comparison, we evaluate four network structures, namely PSPNet \cite{zhao2017pyramid} with ResNet-50 \cite{he2016deep}, DeepLabv3+ \cite{chen2018encoder} with ResNet-50/101, and DeepLabv2 \cite{chen2017deeplab} with ResNet-101. The DeepLabv2 model is initialized with the MS COCO \cite{lin2014microsoft} pre-trained parameters following \cite{ke2020guided}, while the other backbones are pre-trained on ImageNet \cite{deng2009imagenet}.

\vspace{0.1cm}
\noindent
\textbf{Implementation details.} We maintain the same hyper-parameters between supervised pre-training of the teacher and semi-supervised re-training of the student. Specifically, the batch size is set as 16 for both Pascal and Cityscapes with two and four NVIDIA V100 GPUs respectively (actually, two 2080Ti GPUs are already enough for all experiments on Pascal). We use the SGD optimizer for training, where the initial base learning rate of the backbones is set as 0.001 on Pascal and 0.004 on Cityscapes. The learning rate of the randomly initialized segmentation head is 10 times larger than that of backbones. We use the poly scheduling to decay the learning rate during the training process: $lr = baselr \times (1 - \frac{iter}{total\_iter})^{0.9}$. The model is trained for 80 epochs on Pascal and 240 epochs on Cityscapes. For weak data augmentations, the training images are randomly flipped and resized between 0.5 and 2.0. They are further cropped to 321x321 on Pascal and 721x721 on Cityscapes. For the strong data augmentations on unlabeled images, we use colorjitter with the same intensity as \cite{zou2020pseudoseg}, grayscale, blur same as \cite{chen2020improved}, and Cutout with random values filled. The Cutout regions are ignored in loss computation. In the pseudo labeling phase, all unlabeled images are predicted with test-time augmentation, which contains five scales and horizontal flipping. The testing images are evaluated on their original resolution and no post-processing techniques are adopted. The reliable images are measured with three checkpoints that are evenly saved at 1/3, 2/3, 3/3 total iterations during training. We simply treat the top 50\% highest scored images as reliable ones and the remaining ones as unreliable. It is worth noting that, for fair comparison with most existing works, we do not incorporate any advanced optimization strategies, such as OHEM in \cite{hu2021semi}, auxiliary supervision in \cite{chen2021semi, hu2021semi}, nor SyncBN into our method. We also choose a relatively smaller cropping size during training to save memory, compared with CPS \cite{chen2021semi} (321 \textit{vs.}~512 on Pascal and 721 \textit{vs.}~800 on Cityscapes).

\begin{table}
    \centering
    \small
    \begin{tabular}{lccc}
    \toprule
    \multirow{2}*{Method} & \multicolumn{3}{c}{ResNet-50 / ResNet-101} \\
    \specialrule{0pt}{0.5pt}{0.5pt}
    \cmidrule{2-4}
    \specialrule{0pt}{0pt}{0pt}
    ~ & 1/16 (662) & 1/8 (1323) & 1/4 (2645) \\
    \specialrule{0pt}{0.5pt}{0.5pt}
    \midrule
    
    SupOnly & 64.0 / 68.4 & 69.0 / 73.3 & 71.7 / 74.7 \\
    
    CCT \cite{ouali2020semi} & 65.2 / 67.9 & 70.9 / 73.0 & 73.4 / 76.2 \\
    
    CutMix-Seg \cite{french2019semi} & 68.9 / 72.6 & 70.7 / 72.7 & 72.5 / 74.3 \\
    
    GCT \cite{ke2020guided} & 64.1 / 69.8 & 70.5 / 73.3 & 73.5 / 75.3 \\
    
    CPS \cite{chen2021semi} & 68.2 / 72.2 & 73.2 / 75.8  & 74.2 / 77.6 \\
    
    CPS$^\dag$ \cite{chen2021semi} & 72.0 / \textbf{\textit{74.5}} & 73.7 / 76.4 & 74.9 / \textbf{\textit{77.7}} \\

    \midrule
    \textbf{ST} & \textbf{\textit{72.2}} / 74.0 & \textbf{\textit{74.8}} / \textbf{\textit{76.9}} & \textbf{\textit{75.5}} / 77.6 \\
    
    \textbf{ST++} & \textbf{73.2} / \textbf{74.7} & \textbf{75.5} / \textbf{77.9} & \textbf{76.0} / \textbf{77.9} \\
    
    \bottomrule
    \end{tabular}
    \caption{Results on Pascal VOC using a modified ResNet with the deep stem block, following CPS \cite{chen2021semi}. $\dag$: CPS also adopts CutMix \cite{yun2019cutmix} to further boost the performance.}
    \vspace{-0.03cm}
    \label{tab:compare_cps}
\end{table}

\subsection{Comparison with State-of-the-Art Methods}
Two frameworks based on classical self-training scheme are proposed in this work, namely ST and ST++. In this section, we extensively compare both of our frameworks with previous methods across a variety of datasets and settings.

\vspace{0.1cm}
\noindent
\textbf{Pascal VOC 2012.} Most previous works uniformly sample labeled images from the augmented training set, which contains 10,582 images in total.
As shown in Table \ref{tab:pascal_augmented_sota}, our proposed ST already outperforms existing methods remarkably with the four network architectures across extensive settings. Moreover, with the advanced ST++, the performance is further boosted consistently. Besides, the significant margin between our ST/ST++ and the supervised only (SupOnly) results proves our successful and effective exploitation on the abundant unlabeled images. Among most recent methods, CPS \cite{chen2021semi} adopts a stronger ResNet with the deep stem block, therefore we also modify our backbone to conduct a fair comparison in Table \ref{tab:compare_cps}.

Some recent works sample labeled images from the high-quality original training set. We evaluate our method under this setting in Table \ref{tab:pascal_original_sota}. Without iterative training adopted as \cite{yuan2021simple}, our ST and ST++ framework surpass the previous state-of-the-art methods impressively, even outperform the fully-supervised setting with only 1464 labeled images.

\begin{figure}[t]
    \centering
    \includegraphics[width=0.95\linewidth]{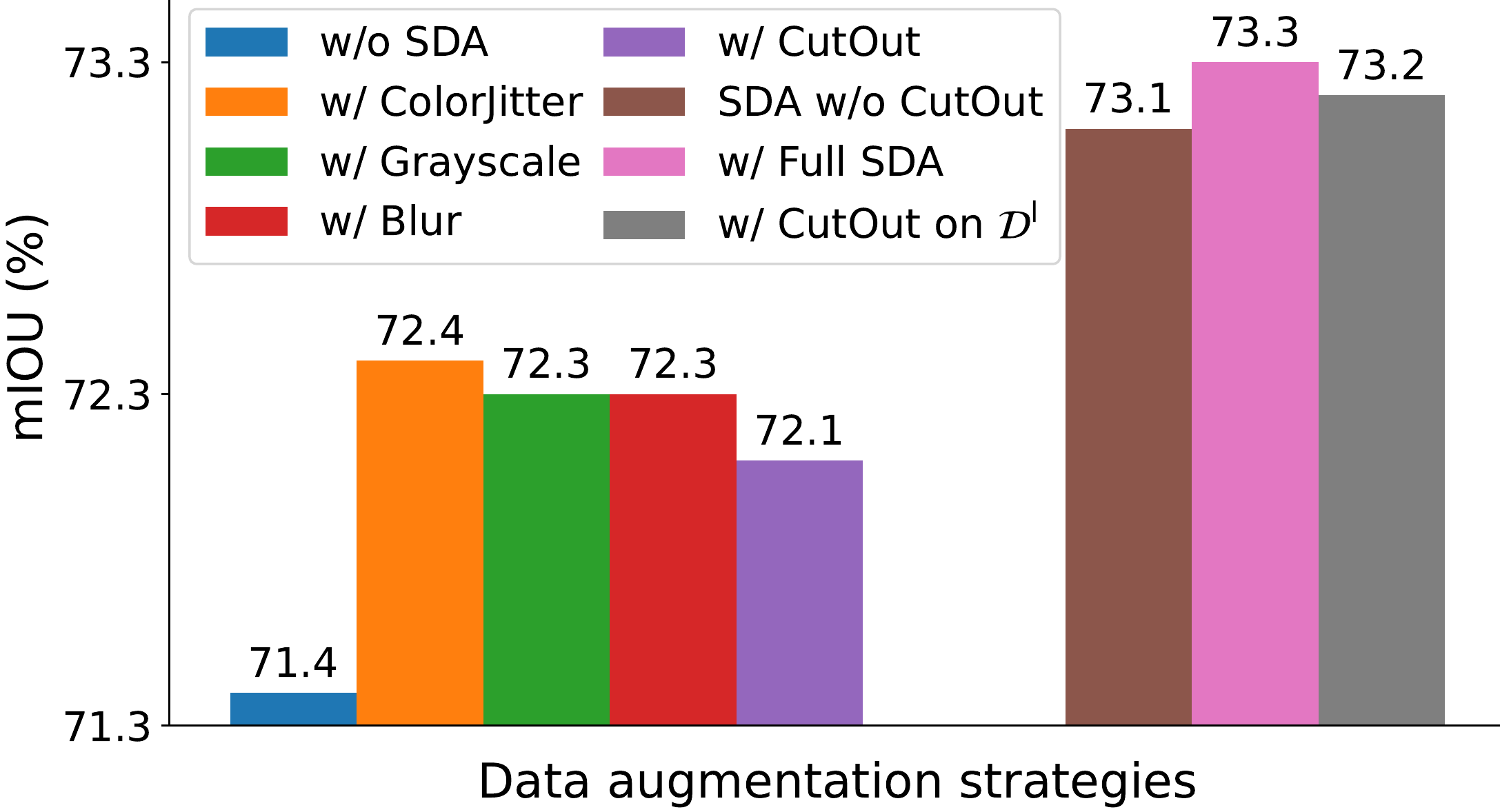}
    \caption{Effectiveness of each data augmentation. All strong data augmentations contributes to the final success. The inferior result of further applying Cutout on labeled images (73.2 \emph{vs}. 73.3) demonstrates that Cutout in SDA works as a strong regularization for unlabeled data rather than a common trick for training.}
    \vspace{-0.03cm}
    \label{fig:each_aug}
\end{figure}

\begin{table}
\centering
\small
\begin{tabular}{lccccc}
    \toprule
    
    \multirow{2}*{Method} & \multicolumn{5}{c}{\# Labeled images (Total: 10582)} \\
    \specialrule{0pt}{0.5pt}{0.5pt}
    \cmidrule{2-6}
    \specialrule{0pt}{0pt}{0pt}
    ~ & 92 & 183 & 366 & 732 & 1464 \\
    \specialrule{0pt}{0.5pt}{0.5pt}
    \midrule
    
    SupOnly & 50.7 & 59.1 & 65.0 & 70.6 & 74.1 \\
    
    GCT \cite{ke2020guided} & 46.0 & 55.0 & 64.7 & 70.7 & - \\ 
    
    CutMix-Seg \cite{french2019semi} & 55.6 & 63.2 & 68.4 & 69.8 & - \\
    
    PseudoSeg \cite{zou2020pseudoseg} & 57.6 & 65.5 & 69.1 & 72.4 & 73.2\\
    
    CPS \cite{chen2021semi} & \textbf{\textit{64.1}} & 67.4 & 71.7 & 75.9 & - \\
    
    PC$^2$Seg \cite{zhong2021pixel} & 57.0 & 66.3 & 69.8 & 73.1 & 74.2 \\
    
    \midrule
    
    \textbf{ST} & 61.3 & \textbf{\textit{68.2}} & \textbf{\textit{73.5}} & \textbf{\textit{76.3}} & \textbf{78.9} \\
    
    \textbf{ST++} & \textbf{65.2} & \textbf{71.0} & \textbf{74.6} & \textbf{77.3} & \textbf{79.1} \\
    \midrule
    
    \multicolumn{6}{c}{\emph{Fully-supervised setting (10582 images): 78.2}} \\
    
    \bottomrule
\end{tabular}
\caption{Results on Pascal VOC using DeepLabv3+ with ResNet-101. Labeled images are selected from \textbf{original} training set.}
\vspace{0.15cm}
\label{tab:pascal_original_sota}
\end{table}

\vspace{0.1cm}
\noindent
\textbf{Cityscapes.} As shown in Table \ref{tab:cityscapes_sota}, across a wide range of the number of labeled images, \eg, from 744 to mere 100, both of our methods obtain the state-of-the-art results under a fair comparison with previous methods. It is worth noting that, our method with ResNet-50 backbone even surpasses other works with ResNet-101 backbone by a large margin.

\vspace{0.1cm}
\noindent
\textbf{Discussion with previous works.} \cite{french2019semi} explores the strong, varied perturbations in the semi-supervised semantic segmentation and finds that Cutout and CutMix play an important role in the consistency regularization. Besides, \cite{zou2020pseudoseg} inherits the spirit from FixMatch \cite{sohn2020fixmatch} and applies strong-weak perturbations for consistency regularization. Different from these one-stage works, we find that a simple offline self-training scheme produces more stable and consistent artificial masks. Moreover, coupled with strong data augmentations which are uncommon in supervised scenarios, \eg, colorjitter, grayscale, and blur on unlabeled data, the offline two-stage pipeline can enforce the consistency across various strong perturbations in a broader stage-wise scope, without the limitation from current minibatch. And empirically, according to extensive validations, our proposed ST surpasses all the prior methods impressively.

\begin{figure}[t]
    \centering
    \includegraphics[width=0.95\linewidth]{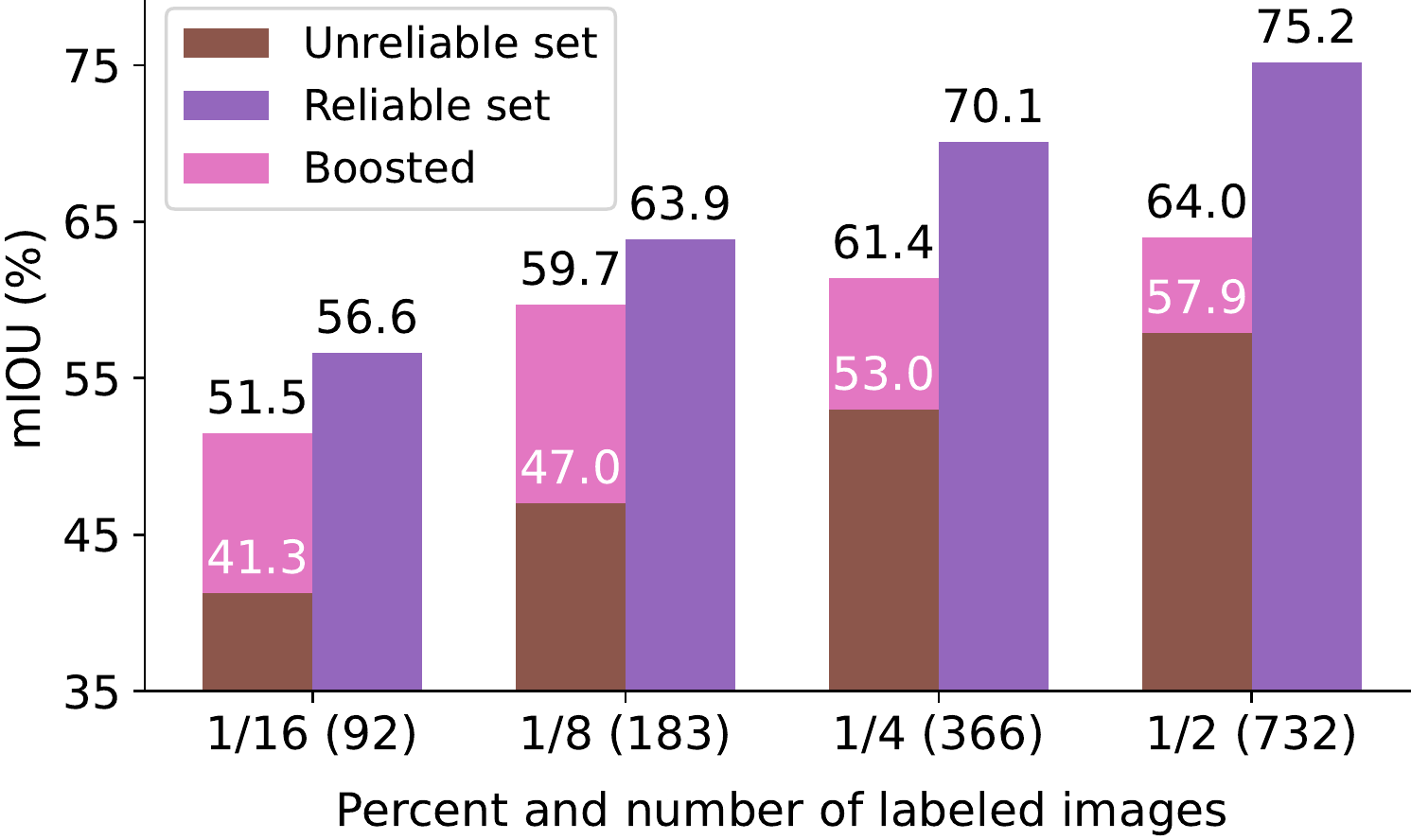}
    \caption{Pseudo mask quality of the reliable and unreliable images selected by ST++. The \textit{Boosted} means the improved mIOU when re-labeling the unreliable images with the model trained on reliable images compared with only trained with labeled images.}
    \vspace{-0.03cm}
    \label{fig:mask_quality}
\end{figure}

\begin{table}
\centering
\small
\begin{tabular}[t]{lccc}
    \toprule
    
    \multirow{1}*{Method} & 1/30 (100) & 1/8 (372) & 1/4 (744) \\
    
    \midrule
    \multicolumn{4}{c}{DeepLabv3+, ResNet-101} \\
    \midrule
    
    DMT \cite{feng2020dmt} & 54.8 & 63.0 & - \\
    
    CutMix-Seg \cite{french2019semi} & 55.7 & 65.8 & 68.3\\
    
    ClassMix \cite{olsson2021classmix} & - & 61.4 & 63.6 \\
    
    PseudoSeg \cite{zou2020pseudoseg} & 61.0 & 69.8 & 72.4\\
    
    \midrule
    \multicolumn{4}{c}{DeepLabv3+, ResNet-50} \\
    \midrule
    
    SupOnly & 55.1 & 65.8 & 68.4 \\
    
    DCC \cite{lai2021semi} & - & 69.7 & 72.7 \\
    
    \textbf{ST} & \textbf{\textit{60.9}} & \textbf{\textit{71.6}} & \textbf{\textit{73.4}} \\
    
    \textbf{ST++} & \textbf{61.4} & \textbf{72.7} & \textbf{73.8} \\
    
    \bottomrule
\end{tabular}
\caption{Results on Cityscapes. It is worth noting that our method with ResNet-50 already surpasses others with ResNet-101.}
\vspace{0.15cm}
\label{tab:cityscapes_sota}
\end{table}

\subsection{Ablation Studies}

The main findings and contributions of this work lie in 1) strong data augmentations (SDA) on unlabeled data and 2) selective re-training. In this section, we examine the actual effectiveness of the two components in detail. We conduct our ablation studies with DeepLabv3+ and ResNet-50 on the Pascal VOC. Unless otherwise specified, 1323 (1/8) labeled images are sampled from the augmented training set.

\vspace{0.1cm}
\noindent
\textbf{Effectiveness of the SDA in ST.} As aforementioned, SDA is composed of colorjitter, grayscale, blur, and Cutout. Here we validate their effectiveness from three perspectives. (1) Firstly, we show the results when no SDA is adopted in Table \ref{tab:ablation_sda}. It can be seen that the model performance degrades across all settings, proving that SDA is vital to the success of self-training. (2) Following this, we further examine the effect of SDA on labeled images. According to the results in Table \ref{tab:ablation_sda}, the labeled images are negatively affected by SDA, indicating that it may deteriorate the clean distribution of labeled images. (3) Finally, in order to gain a better intuition of the individual benefit of the four data augmentations, we add each of them to unlabeled images in Figure \ref{fig:each_aug}. 

Besides these, since Cutout is proposed as a common trick in image recognition, we attempt to also apply it to labeled images (gray column in Figure \ref{fig:each_aug}). The inferior results prove that our proposed ST benefits from the strong perturbations on unlabeled images rather than common tricks.

\begin{figure}[t]
    \centering
    \includegraphics[width=0.95\linewidth]{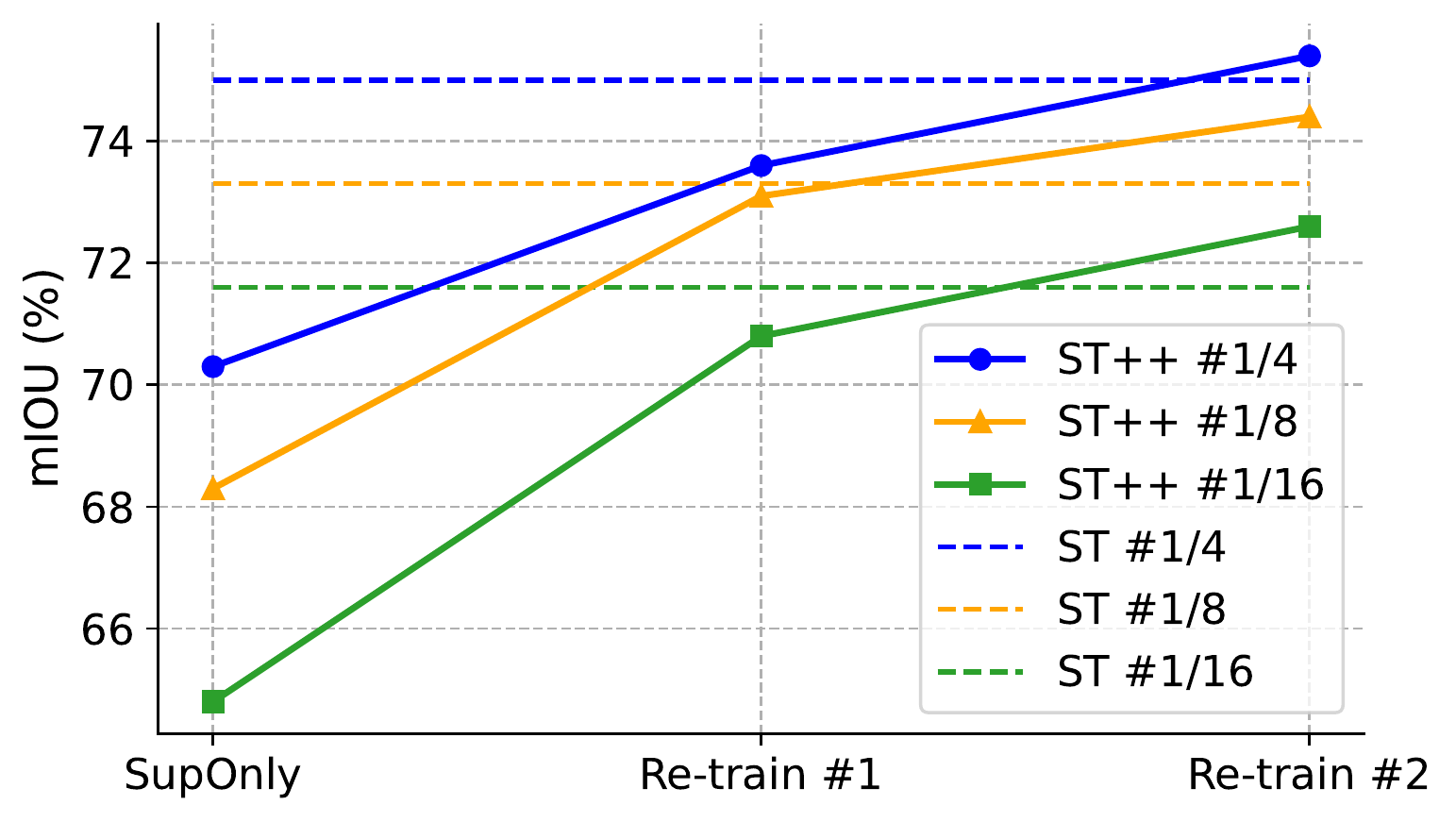}
    \caption{Training process of our proposed ST++. ST++ first trains the model on reliable images (Re-train \#1) and then on all re-labeled images (Re-train \#2). It can be seen the first stage re-training already obtains competitive results as our ST, further proving that the image-level selection is appropriate and necessary.}
    \vspace{-0.03cm}
    \label{fig:st++_progess}
\end{figure}

\begin{table}[t!]
\centering
\small
\begin{tabular}{ccccc}
    \toprule
    
    \multicolumn{2}{c}{Apply SDA on} & 1/16 & 1/8 & 1/4 \\
    labeled data & unlabeled data & (662) & (1323) & (2645) \\
    
    \midrule
    
    ~ & ~ & 70.9 & 71.4 & 73.5 \\
    
    \cmark & \cmark & 71.0 & 73.0 & 74.3 \\
    
    ~ & \cmark & \textbf{71.6} & \textbf{73.3} & \textbf{75.0} \\
    
    \bottomrule
\end{tabular}
\caption{Effectiveness of full SDA. The first line without applying SDA is the plainest self-training \cite{lee2013pseudo}. The best results of only applying SDA on unlabeled data indicates that a more challenging optimization target for unlabeled data is vital to the success. And SDA on labeled data may destroy the clean data distribution.}
\vspace{0.15cm}
\label{tab:ablation_sda}
\end{table}

\vspace{0.1cm}
\noindent
\textbf{Effectiveness of selective re-training in ST++.} We first measure the quality of the pseudo masks from reliable and unreliable set respectively. As shown in Figure \ref{fig:mask_quality}, the meanIOU gap between pseudo masks from reliable set and unreliable set is all larger than 15\%, indicating that it is reasonable to select and prioritize the reliable set. The obtained better student can produce more accurate pseudo masks on the remaining unreliable images. The improvement of unreliable image masks is remarkable as the \textit{Boosted} column.

We further check whether our pipeline benefits from the correct selection of reliable unlabeled images or merely the two-stage training pipeline. We randomly divide the whole unlabeled training set into two parts, training with one part first and then re-labeling the remaining ones. The model is finally jointly optimized on the full combination of manually labeled and pseudo labeled images. As shown in Table \ref{tab:ablation_select_retrain}, the semi-supervised model does not benefit from the random two-stage training, in some cases even inferior to its one-stage counterpart. As a comparison, our selective re-training based on image-level stability and reliability consistently outperforms the one-stage re-training pipeline.

Since the re-training phase is conducted in a two-stage manner in our ST++, we examine the improvement after each stage in Figure \ref{fig:st++_progess}. It can be easily seen that after the first stage, where only half the unlabeled images are exploited, the ST++ already obtains competitive results as ST, revealing the high quality of the selected reliable images.

We also conduct ablation studies on the proportion of selected reliable images. As shown in Table \ref{tab:reliable_proportion}, the default hyper-parameter 50\% is effective enough. Our ST++ is also robust to other values and the 75\% is even slightly better.

\begin{table}[t!]
    \centering
    \small
    \begin{tabular}{lccc}
    
    \toprule
    
    \multirow{2}*{Method} & 1/16 & 1/8 & 1/4 \\
    ~ & (662) & (1323) & (2645) \\
    
    \midrule
    
    One-stage re-training (our ST) & 71.6 & 73.3 & 75.0 \\
    
    Random two-stage re-training & 71.3 & 73.9 & 74.7 \\
    
    Selective re-training (our ST++) & \textbf{72.6} & \textbf{74.4} & \textbf{75.4} \\
    
    \bottomrule
     
    \end{tabular}
    \caption{Effectiveness of the selective re-training in ST++. ST++ does not benefit from random two-stage re-training process, but the progressive reliable-to-unreliable selective re-training pipeline.}
    \vspace{-0.03cm}
    \label{tab:ablation_select_retrain}
\end{table}

\begin{table}[t]
    \centering
    \small
    \begin{tabular}{lccc}
    
    \toprule
    
    Proportion of reliable images & 25\% & 50\% (default) & 75\% \\
    
    \midrule
    
    mIOU (\%) & 74.0 & 74.4 & 74.5 \\
    
    \bottomrule
     
    \end{tabular}
    \caption{Ablation study on the proportion of reliable images. The default hyper-parameter 50\% is effective enough.}
    \vspace{0.15cm}
    \label{tab:reliable_proportion}
\end{table}

\vspace{0.1cm}
\noindent
\textbf{Comparison between image-level and pixel-level selective re-training.} The contribution in our ST++ is image-level selective re-training, where we decompose the re-training phase into two sub-phases and prioritize reliable images for the first phase re-training. We argue that image-level selection is more stable and provides complete context information for training. Therefore, we compare it with pixel-level selection, where high-confidence pixels are selected for the first phase re-training, and the remaining pixels are re-labeled with a better student for the second phase re-training. Following \cite{zoph2020rethinking, zou2020pseudoseg}, we set the confident threshold as 0.5. As demonstrated in Table \ref{tab:pixel_retrain}, the image-level selection brings consistent improvements over our ST framework and is superior to the pixel-level counterpart.

\begin{figure}[t]
    \centering
    \includegraphics[width=0.95\linewidth]{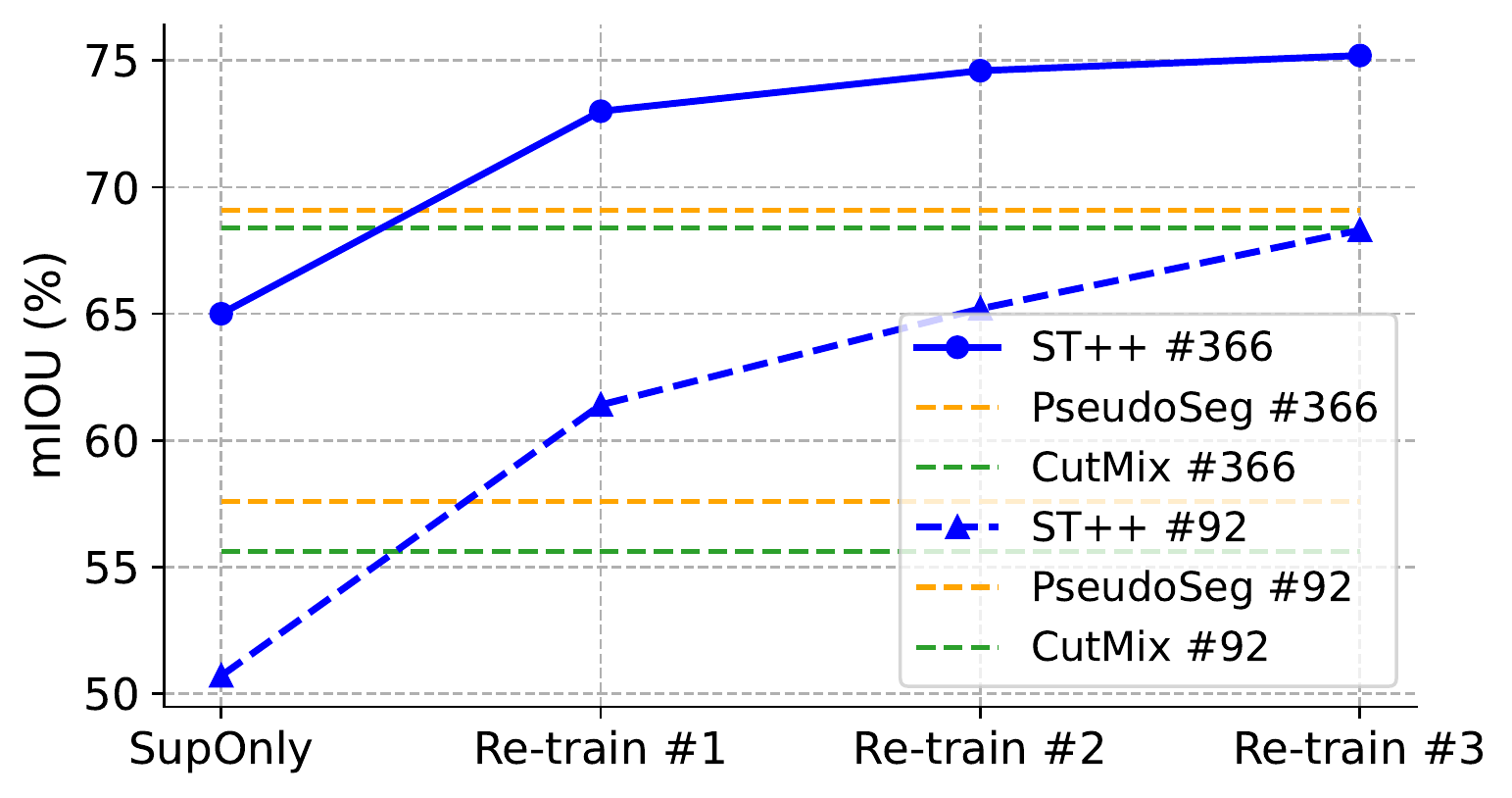}
    \caption{Effectiveness of iterative training. We also show results of PseudoSeg \cite{zou2020pseudoseg} and CutMix-Seg \cite{french2019semi} for a clear comparison.}
    \vspace{-0.03cm}
    \label{fig:iterative}
\end{figure}

\begin{table}[t!]
    \centering
    \small
    \begin{tabular}{lccc}
    
    \toprule
    
    \multirow{2}*{Method} & 1/16 & 1/8 & 1/4 \\
    
    ~ & (662) & (1323) & (2645) \\
    
    \midrule
    
    our ST (w/o iterative re-train) & 71.6 & 73.3 & 75.0 \\
    
    Image-level re-train (our ST++) & \textbf{72.6} & \textbf{74.4} & \textbf{75.4} \\
    
    \midrule
    
    Pixel-level re-train phase \#1 & 71.3 & 73.5 & 74.9 \\
    
    Pixel-level re-train phase \#2 & 71.3 & 73.8 & 74.7 \\
    
    \bottomrule
     
    \end{tabular}
    \caption{Comparison between image-level (our ST++) and common pixel-level selective re-training (setting a threshold).}
    \vspace{0.15cm}
    \label{tab:pixel_retrain}
\end{table}

\vspace{0.1cm}
\noindent
\textbf{Effectiveness of iterative training.} As aforementioned, for simplicity and efficiency, we choose not to conduct iterative training. However, it is possible to further boost the final performance via switching the teacher-student role and re-labeling unlabeled images for several times. Therefore, we examine the effectiveness of iterative training in our ST++ in Figure \ref{fig:iterative}. An extra stage (Re-train \#3 in the figure) is conducted, where all the unlabeled images are re-labeled with the best learned model in the second re-training stage and the student is re-trained. In the extremely scarce-label regime of only 92 and 366 labeled images, with the extra re-training stage, the performance can be further boosted from 65.2\% to 68.3\% and 74.6\% to 75.2\% respectively.

\section{Conclusion}

In this work, we firstly construct a strong self-training baseline for semi-supervised semantic segmentation via introducing strong data augmentations to unlabeled images, in hope of alleviating overfitting noisy labels as well as decoupling similar predictions between the teacher and student. Moreover, an advanced framework is proposed to progressively leverage the unlabeled images. With extensive experiments conducted across a variety of benchmarks and settings, both of our ST and ST++ framework outperform previous methods by a large margin. Based on the inspiring results, we further examine the effectiveness of each component in detail and provide some empirical analysis. We hope this simple yet effective framework can serve as a strong baseline or competitor for future works in this field.

\clearpage
{\small
\bibliographystyle{ieee_fullname}
\bibliography{egbib}
}

\end{document}